\documentclass[11pt]{article}
\usepackage{coling2020}
\usepackage{times}
\usepackage{multirow,booktabs}
\usepackage{capt-of}
\usepackage[table]{xcolor}
\usepackage{url}
\usepackage{latexsym}
\usepackage{fdsymbol}
\usepackage{bm}
\usepackage{todonotes}
\usepackage{microtype}
\usepackage{scrextend}
\usepackage{ctable}
\usepackage{array}

\definecolor{hardyred}{HTML}{990000}
\definecolor{darkgreen}{HTML}{006330}

\newcolumntype{P}[1]{>{\centering\arraybackslash}p{#1}}

\colingfinalcopy 

\title{Diverse Keyphrase Generation with Neural Unlikelihood Training}

\author{Hareesh Bahuleyan\footnotemark[2]\quad  Layla El Asri\footnotemark[2]\\
  \footnotemark[2]\hspace{0.2em} Borealis AI, Montreal, Canada  \\
  {\tt layla.elasri@borealisai.com}\\ {\tt hareeshbahuleyan@gmail.com}\\ }

\date{}

\begin{document}
\maketitle
\begin{abstract}
In this paper, we study sequence-to-sequence (S2S) keyphrase generation models from the perspective of diversity. Recent advances in neural natural language generation have made possible remarkable progress on the task of keyphrase generation, demonstrated through improvements on quality metrics such as $F_1$-score. However, the importance of diversity in keyphrase generation has been largely ignored. We first analyze the extent of information redundancy present in the outputs generated by a baseline model trained using maximum likelihood estimation (MLE). Our findings show that repetition of keyphrases is a major issue with MLE training. To alleviate this issue, we adopt neural unlikelihood (UL) objective for training the S2S model. Our version of UL training operates at (1) the target token level to discourage the generation of repeating tokens; (2) the copy token level to avoid copying repetitive tokens from the source text. Further, to encourage better model planning during the decoding process, we incorporate $K$-step ahead token prediction objective that computes both MLE and UL losses on future tokens as well. Through extensive experiments on datasets from three different domains we demonstrate that the proposed approach attains considerably large diversity gains, while maintaining competitive output quality.\footnote{Code is available at \url{https://github.com/BorealisAI/keyphrase-generation}}
\end{abstract}

\section{Introduction}\blfootnote{This work is licensed under a Creative Commons Attribution 4.0 International License. License
details: \url{http://creativecommons.org/licenses/by/4.0}/}
Automatic keyphrase generation is the task of generating single or multi-word lexical units that provides readers with high level information about the key ideas or important topics described in a given source text. Apart from an information summarization perspective, this task has applications in various downstream natural language processing tasks such as text classification \cite{liu-etal-2009-unsupervised}, document clustering \cite{hammouda2005corephrase} and information retrieval \cite{nguyen2007keyphrase}. 

Traditionally, keyphrases (KPs) were \textit{extracted} from source documents by retrieving and ranking a set of candidate phrases through rule based approaches. With recent advances in neural natural language generation and availability of larger training corpora, this problem is formulated under a sequence-to-sequence (S2S) modelling framework \cite{sutskever2014sequence}. This approach has an advantage that it can \textit{generate} new and meaningful keyphrases which may be absent in the source text. The earliest work in this direction was by \newcite{meng-deep-kp}, who train a S2S model to generate one keyphrase at a time. At inference time, they decode with beam sizes as high as 200, to generate a large number of KPs and finally de-duplicate the outputs. However, this is computationally expensive and wasteful because only $<5\%$ of such KPs were found to be unique \cite{yuan-catseqkp}. 

An alternative approach is to train a S2S model to generate multiple keyphrases in a sequential manner, where the output KPs are separated by a pre-defined delimiter token. This method has an added benefit that the model automatically learns to generate a variable number of keyphrases depending on the input, instead of a user-specified fixed number of keyphrases (top-$k$) from a large list of candidate outputs. However, some previous approaches \cite{yuan-catseqkp} still use exhaustive beam search decoding to over-generate KPs and then apply post-processing to remove repetitions. Apart from the additional computational requirements, we argue that this method of  avoiding information redundancy is a last-minute solution. 


In this paper, we take a principled direction towards addressing the information redundancy issue in keyphrase generation models. We propose to tackle this problem directly during the training stage, rather than applying adhoc post-processing at inference time. Specifically, we adopt the neural unlikelihood training (UL) objective \cite{welleck-ul}, whereby the decoder is penalized for generating undesirable tokens.
\newcite{welleck-ul} introduce unlikelihood training for a language model setting. Since we work with a S2S setup, our version of UL loss consists of two components: (1) a target token level UL loss based on the target vocabulary to penalize the model for \textit{generating} repeating tokens; (2) a copy token level UL loss based on the dynamic vocabulary of source tokens required for copy mechanism \cite{copymech,see-pointer}, which penalizes the model for \textit{copying} repetitive tokens. 

S2S models trained with maximum likelihood estimation (MLE) are usually tasked with the next token prediction objective. However, this does not necessarily incentivize the model to plan for future token prediction ahead of time. We observe such lack of model planning capability in our initial experiments with MLE models and to overcome this issue we propose to use $K$-step ahead token prediction. This modified training objective encourages the model to learn to correctly predict not just the current token, but also tokens upto $K$-steps ahead in the future. We then naturally incorporate UL training on the $K$-step ahead token prediction task.

We summarize our contributions as follows: (1) To improve the diversity of generated keyphrases in a principled manner during training, we adopt the unlikelihood objective for the S2S setting and propose a novel copy token unlikelihood loss. (2) In order to incentivize model planning, we augment our training objective function to incorporate $K$-step ahead token prediction. Additionally, we also introduce the $K$-step ahead unlikelihood losses. (3) We propose new metrics for benchmarking keyphrase generation models on diversity criterion. We carry out experiments on datasets from three different domains (scientific articles, news and community QA) and validate the effectiveness of our approach.  We observe substantial gains in diversity while maintaining competitive output quality.

\begin{table}[!t]
\small
\centering
    \begin{tabular}{|p{6.7em}|p{40.9em}|}
    \toprule
    \textbf{Title} & semi automated schema integration with sasmint  \\
    \midrule
    \textbf{Abstract} & the emergence of increasing number of collaborating organizations has made clear the need for supporting interoperability infrastructures , enabling sharing and exchange of data among organizations . schema matching and schema integration are the crucial components of the interoperability infrastructures , and their semi automation to interrelate or integrate heterogeneous and autonomous databases in collaborative networks is desired . the semi automatic schema matching and integration sasmint system introduced in this paper identifies and resolves (...)  \\
    \midrule
    \textbf{Ground Truth} & \textit{schema integration ; collaboration ; schema matching ; heterogeneity ; data sharing} \\
    \midrule
    \textbf{MLE Baseline} & \textit{\textcolor{red}{schema integration} ; sasmint ; \textcolor{red}{schema matching} ; \textcolor{red}{schema integration} ; \textcolor{red}{schema matching} ; sasmint derivation markup language
    } \\
    \midrule
    \textbf{DivKGen} & \textit{schema integration ; interoperability infrastructures ; schema matching ; sasmint} \\
    \bottomrule
    \end{tabular}
  \label{tab:example_diversity}
  \vspace{-0.7em}
  \caption{Comparison of sample outputs generated by our model (DivKGen) vs. an MLE baseline. The repeating keyphrases are shown in \textit{\textcolor{red}{red}}.}
\end{table}

\section{Background and Motivation}
\subsection{Problem Definition}
The task of keyphrase generation can be formulated in the following manner. Given a source document $\mathbf{x}$, we are required to generate a set of keyphrases $\mathcal{Y}=\{\mathbf{y}^1, \mathbf{y}^2, \ldots, \mathbf{y}^{|\mathcal{Y}|}\}$ that best describe the input. The source document is denoted as a sequence of $S$ words: $\mathbf{x} = (x_1, x_2, \ldots, x_{S})$. Each target keyphrase $\mathbf{y}^i=(y_1, y_2,\ldots,y_{T_i})$ is also a word sequence of length $T_i$.

We follow the modelling setup adopted in previous work on keyphrase generation~\cite{semi-supervised-kp,reinforce-kp}. Given document-keyphrases pair $(\mathbf{x}, \mathcal{Y})$, we concatenate all the ground truth keyphrases into a single linearized output sequence $\mathbf{y} = \mathbf{y}^1 \smalldiamond \mathbf{y}^2 \smalldiamond \ldots \smalldiamond \mathbf{y}^{|\mathcal{Y}|}$, where $\smalldiamond$ denotes a special delimiter token that is inserted in between consecutive keyphrases. The training data now consists of $(\mathbf{x}, \mathbf{y})$ pairs and one can conveniently use a sequence-to-sequence (S2S) modelling architecture to learn the mapping from $\mathbf{x}$ to $\mathbf{y}$.

\subsection{Sequence Encoder-Decoder}
A bi-directional LSTM encoder \cite{lstm} reads the variable length source sequence  $\mathbf{x} = (x_1, \ldots, x_i, \ldots, x_{S})$ and produces a sequence of hidden state representations $h = (h_1, \ldots, h_i, \ldots, h_{S})$ with $h_i \in \mathbb{R}^{d_h}$,  using the operation $h_i = f_{enc}(x_i, h_{i-1})$ where $f_{enc}$ is a differentiable non-linear function. 

For the decoder, we use a uni-directional LSTM which computes a hidden state $s_t \in \mathbb{R}^{d_s}$ at each decoding time step based on a non-linear function defined as $s_t = f_{dec}(e_{t-1}, s_{t-1})$. At training time, $e_{t-1}$ is the embedding of the ground truth previous word and at inference time, it is the embedding of the word 
predicted at the previous time step.

\subsection{Attention Guided Decoding}
By incorporating global attention mechanism \cite{bahdanau-attn} into the basic S2S architecture, it is possible to dynamically align source information with the target hidden states during the decoding process. This is achieved by computing an alignment score between the decoder hidden state $s_t$ and each of the encoder hidden representations $\{\bm h_i\}_{i=1}^{S}$. At decoding time step $t$, this corresponds to
\begin{equation}
    \alpha_{ti}=\frac{\exp\{\widetilde{\alpha}_{ti}\}}{\sum_{i'=1}^{S}\exp\{\widetilde{\alpha}_{ti'}\}} \enspace ; \enspace \mathrm{where} \enspace
    \widetilde{\alpha}_{ti}= s_t \mathbf{W}_a h_i
    \label{eqn:attn-unnormalized}
\end{equation}
where $\alpha_{ti}$ is referred to as the attention probability score and $\mathbf{W}_a$ is a learnable attention weight matrix. Next we compute the attention context vector as a weighted summation across source hidden states.
\begin{equation}
c_t = \sum_{i=1}^{S} \alpha_{ti} h_i
\label{eqn:attn-ctx}
\end{equation}

\noindent Finally, the probability distribution over a predefined vocabulary $\mathcal{V}_{\mathrm{Target}}$ of target tokens is obtained as
\begin{equation}
    P_{target}(y_t) = \operatorname{softmax}(\mathbf{W}_v\widetilde{s}_t) \enspace ; \enspace \mathrm{where} \enspace
    \widetilde{s}_t = \operatorname{tanh}(\mathbf{W}_u\left[s_t \oplus c_t\right])
    \label{eqn:output-transformation}
\end{equation}
where $\oplus$ refers to the concatenation operator. Note that $\mathbf{W}_u$ and $\mathbf{W}_v$ are trainable decoder parameters and $y_t \in \mathcal{V}_{\mathrm{Target}}$ . For notational brevity, we omit the bias terms.

\subsection{Copy Mechanism}
\label{sec:copying}
We incorporate copy mechanism \cite{copymech} to alleviate the out-of-vocabulary issue during generation, by allowing the decoder to selectively \textit{copy} tokens from the source document. Specifically, we employ a learnable switching parameter $p_{gen} = \operatorname{sigmoid}(\mathbf{W}_c\left[s_t; c_t;e_{t-1}\right])$ which refers to the probability of generating a token from the target vocabulary $\mathcal{V}_{\mathrm{Target}}$. Thus, $(1-p_{gen})$ corresponds to the probability of copying a token present on the source side whose dynamic vocabulary is denoted by $\mathcal{V}_{\mathbf{x}}$.
The generation probability and the copy probability at time step $t$ are then combined to predict the next token as follows:
\begin{equation}
    P(y_t) = p_{gen}P_{target}(y_t) + (1-p_{gen})P_{copy}(y_t)
\end{equation}
\noindent where $y_t \in \mathcal{V}_{\mathrm{Target}} \cup \mathcal{V}_{\mathbf{x}}$ and $P_{copy}(y_t) = \sum_{i:x_i=y_t} \alpha_{ti}$ is the copy probability of token $y_t$ defined as a sum of its attention weights across all its occurrences in the source text.

\subsection{Maximum Likelihood Training}
Encoder-decoder models for sequence generation are typically trained using Maximum Likelihood Estimation (MLE). Concretely, for a given instance in the training data, MLE objective corresponds to learning the model parameters $\bm \theta$ that minimizes the negative log-likelihood loss defined as follows: 
\begin{equation}
   \mathcal{L}_{\mathrm{MLE}} = - \sum_{t=1}^L \log{P(y_t|\mathbf{y}_{1:t-1}, \mathbf{x}, \bm \theta)}
   \label{eqn:mle}
\end{equation}
\noindent where $y_t$ is the $t$-th token in the ground truth output sequence $\mathbf{y}$ whose total length is $L$ tokens.

We begin with a setup where the S2S model for keyphrase generation is trained using MLE. We carry out preliminary experiments analyzing the diversity of the generation process and demonstrate the shortcomings of MLE-based training (Section~\ref{sec:lackOfDiversity}) which paves way for the proposed approach (Section~\ref{sec:proposed}). 

\subsection{Lack of Diversity Issue}
\label{sec:lackOfDiversity}
We conduct a pilot study using \texttt{KP20k} dataset \cite{meng-deep-kp}, a corpus of scientific articles. Each article consists of a title, an abstract and a set of associated keyphrases. Table~\ref{tab:example_diversity} shows one such example, along with outputs from two systems - a S2S model trained purely with MLE objective and our proposed model which is trained with a combination of unlikelihood training and future token prediction. It can be observed that with MLE objective alone, the S2S model tends to generate the same keyphrase over and over again. On the other hand, the output keyphrases from the proposed model summarizes the abstract of the scientific article, without any repetitions. 

Furthermore, in Table~\ref{tab:lack_of_diversity} we quantify this lack of diversity issue using two simple metrics - the percentage of duplicate keyphrases and the percentage of duplicate tokens. On average, for an MLE model, about 27\% of the generated KPs and 36\% of the generated tokens are duplicates. These values are much higher than  the percentage of repetitions present in the ground truth data. This implies that a significant computational effort is spent in the generation of redundant information. Moreover, additional post-processing pipelines are required in order to get rid of these repetitions. From a user experience point of view, the developed system should generate high quality KPs that describe the main ideas in the source text, without any information redundancy. We design our system keeping this objective in mind. 

\begin{table}[!t]
\small
  \centering
    \begin{tabular}{c|c|c|c}
    \toprule
          & \textbf{\#Keyphrases} & \textbf{\% duplicate keyphrases} & \textbf{\% duplicate tokens} \\
    \midrule
    \textbf{Ground Truth} & 5.3   & 0.1   & 7.3 \\
    \textbf{MLE Baseline} & 7.3   & 26.6  & 36.0 \\
    \bottomrule
    \end{tabular}%
    \caption{A pilot study on \texttt{KP20k} dataset validates our hypothesis about MLE-based training, which tends to generate a large number of repetitions in its outputs. The reported numbers are obtained by averaging the metrics across the test set. }
    \label{tab:lack_of_diversity}
\end{table}%

\section{Proposed Approach}
\label{sec:proposed}
\begin{figure}
        \centering
        \includegraphics[width=1.0\linewidth]{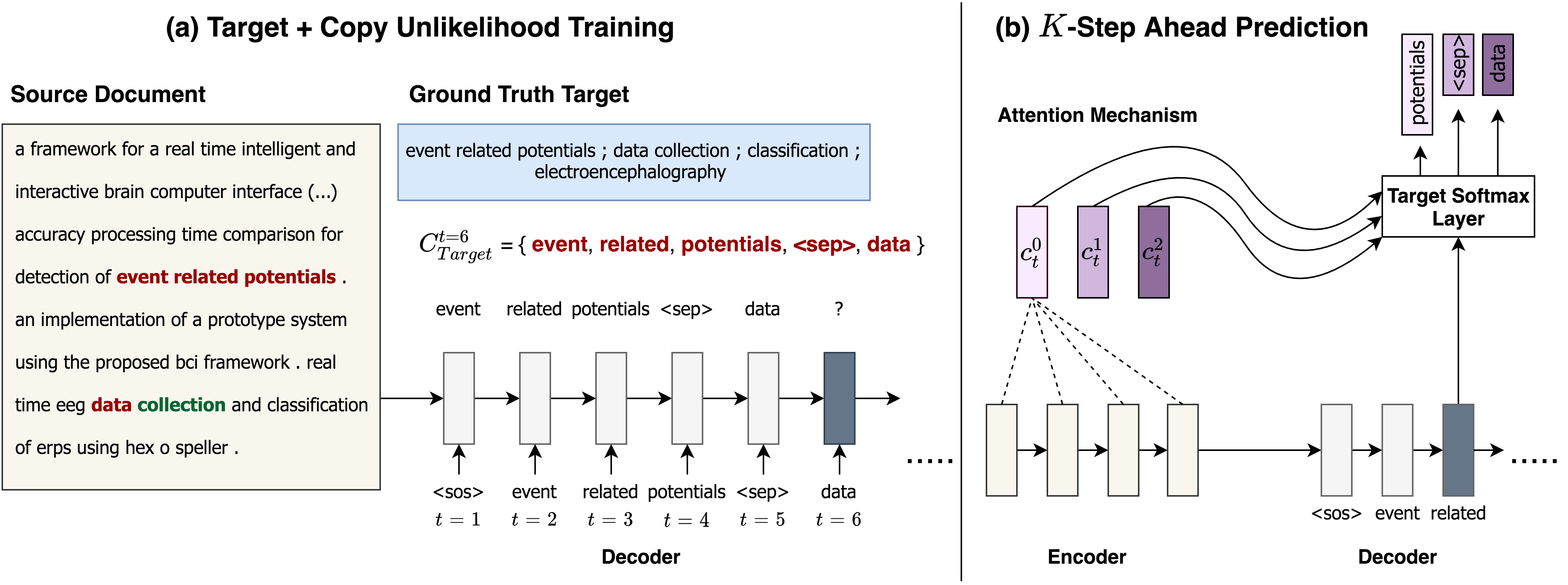}
        \caption{(a) Illustration of unlikelihood training. In the above example, at decoding time step $t=6$, the previous tokens from the target context form the negative candidate list denoted by  $\mathcal{C}_{\mathrm{Target}}^{t=6}$. The Target UL loss is computed using the probabilities assigned to these tokens. Similarly, the Copy UL loss discourages the model from copying the words displayed in \textcolor{hardyred}{\textbf{red}} from the source document at $t=6$. Ideally, we would like the model to copy the word `\textcolor{darkgreen}{\textbf{collection}}'. (b) Depiction of $K-$step ahead prediction with $K=2$. Different attention matrices are used to compute the corresponding attention context vectors for $k=0, 1, 2$. These are then individually fed to the final $\operatorname{softmax}$ layer along with the shared decoder hidden state to predict the token at the respective $k$. Copy mechanism is omitted from Figure 1(b) for simplicity.}
        \label{fig:model}
\end{figure}

Rather than addressing the information redundancy issue through post-processing, we take a principled approach in this direction during training itself. 
Firstly, we adopt neural unlikelihood training \cite{welleck-ul} for sequence-to-sequence setting by directly penalizing the decoder for either generating or copying repeating tokens. 
Secondly, we improve the planning capability of the decoder by incorporating a $K$-step ahead token prediction loss. This is achieved by using the same decoder hidden state but different attention mechanisms to decide which source tokens to be attended to, for predicting the target at the current time step, $1$-step ahead and so on. An illustration of our approach is presented in Figure~\ref{fig:model}.
 
\subsection{Target Token Unlikelihood Loss}
\label{sec:tgt-ul}
The goal of unlikelihood training is to suppress the model's tendency to assign high probabilities to unnecessary tokens. During decoding, say at time step $t$, we maintain a negative candidate list $\mathcal{C}_{\mathrm{Target}}^t$ that consists of tokens that should ideally be assigned a low probability for the current time step prediction. Formally, given $\mathcal{C}_{\mathrm{TargetUL}}^t = \{c_1, \ldots, c_m\}$ where $c_j \in \mathcal{V}_{\mathrm{Target}}$, we define the unlikelihood loss based on the target vocabulary across all time steps as follows
\begin{equation}
    \mathcal{L}_{\mathrm{TargetUL}}=-\sum_{t=1}^L \sum_{\quad c \in \mathcal{C}_{\mathrm{Target}}^{t}} \log \left(1- {P_{target}(c| \mathbf{y}_{1:t-1}, \mathbf{x}, \bm \theta)} \right)
    \label{eqn:tgt-ul}
\end{equation}

\noindent Intuitively, assigning a high probability to a negative candidate token leads to a larger loss. Following \newcite{welleck-ul}, our negative candidate list for $\mathcal{L}_{\mathrm{TargetUL}}$ consists of the ground truth context tokens from the previous time steps, i.e., $\mathcal{C}_{\mathrm{Target}}^t = \{y_1, \ldots , y_{t-1}\} \setminus \{y_t\}$. In this manner, we effectively discourage the model from repeatedly generating tokens that are already present in the previous contexts.  

\subsection{Copy Token Unlikelihood Loss}
\label{sec:copy-ul}
In contrast to \newcite{welleck-ul} who introduce UL training for language model setting, our application employs this method for a S2S task. As described in Section~\ref{sec:copying}, our decoder utilizes a copy mechanism that dynamically creates an extended vocabulary during generation based on the source tokens ($\mathcal{V}_{\mathbf{x}}$). An undesirable side-effect of copying is that the model might repeatedly attend to (and copy) the same set of source tokens over multiple decoding time steps, leading to repetitions in the output. To circumvent this issue, we propose an approach that we refer to as copy token unlikelihood loss. 

For penalizing unnecessary copying, our negative candidate list at each time step is composed of ground truth context tokens from previous time steps that also appear in the source text (and thus can be copied).

\begin{equation}
    \mathcal{L}_{\mathrm{CopyUL}}=-\sum_{t=1}^L \sum_{\quad c \in \mathcal{C}_{\mathrm{Copy}}^{t}} \log \left(1- {P_{copy}(c| \mathbf{y}_{1:t-1}, \mathbf{x}, \bm \theta)} \right)
    \label{eqn:copy-ul}
\end{equation}

\noindent where $\mathcal{C}_{\mathrm{Copy}}^t = \{y_i\ \mid y_i \in \{y_1, \ldots , y_{t-1} \} \setminus \{y_t\} \enspace \mathrm{and} \enspace y_i \in \mathcal{V}_{\mathbf{x}} \}$ and $P_{copy}(c|.)$ refers to the probability of copying a given token $c$ determined by the attention mechanism over the source tokens (Section~\ref{sec:copying}). 


\subsection{$K$-Step Ahead Token Prediction Loss}
\label{sec:future-mle}
Keyphrases are made up of one or more tokens. The decoder in S2S models is usually tasked with simply predicting the next token given the context so far. This greedy approach does not incentivize the model to plan for the upcoming future tokens ahead of time. We mitigate this issue by directly incorporating the prediction of tokens $K$-steps ahead from the current time step into our training objective. 
To do so, we start with Equation~\ref{eqn:mle}, the MLE-based objective for next token prediction at time step $t$. This can be generalized for the prediction of upto $K$ tokens ahead in time as follows: 
\begin{equation}
   \mathcal{L}_{K\mathrm{-StepMLE}} = - \sum_{t=1}^L \sum_{k=0}^K \gamma_k \log{P(y_{t+k}|\mathbf{y}_{1:t-1}, \mathbf{x}, \bm \theta)}
   \label{eqn:kstep-mle}
\end{equation}

\noindent where $\gamma_k$ refers to the coefficient of the $k$th step ahead token prediction loss. Note that the next token prediction MLE objective in Equation~\ref{eqn:mle} is a special case of Equation~\ref{eqn:kstep-mle} where $K=0$ and $\gamma_0=1.0$. One can think of the $K$-step ahead losses as a way to reward the model to plan the surface realization of the output sequence ahead of time. Intuitively, it makes sense to assign a high weightage to current token prediction (i.e., for $k=0$) and relatively downweight the losses incurred from future token predictions. We accomplish this by decaying the coefficient $\gamma_k$ using the formula $\gamma_k = \frac{1.0}{k+1}$.

For $K$-Step ahead prediction, we consider two implementation choices: (1) For each $k$, learn a different transformation $\mathbf{W}_v^k$ (in Equation~\ref{eqn:output-transformation}) from the hidden representation to the logits over the vocabulary $\mathcal{V}_{\mathrm{Target}}$. However, this increases the number of model parameters by $k \times d_{s_t} \times |\mathcal{V}_{\mathrm{Target}}|$ where $d_{s_t}$ is the decoder hidden size. (2) With the second option, for each $k$, a different attention weight matrix $\mathbf{W}_a^k$ is learnt, while having a shared output transformation layer based on $\mathbf{W}_v$. More specifically, Equations~\ref{eqn:attn-unnormalized} and ~\ref{eqn:attn-ctx} can be re-written as
\begin{equation}
    \widetilde{\alpha}^k_{ti}= s_t \mathbf{W}^k_a h_i 
    \quad ; \quad
    \alpha^k_{ti}=\frac{\exp\{\widetilde{\alpha}^k_{ti}\}}{\sum_{i'=1}^{S}\exp\{\widetilde{\alpha}^k_{ti'}\}}
    \quad ; \quad
    c^k_t = \sum_{i=1}^{S} \alpha^k_{ti} h_i
\end{equation}

The intuition behind such a formulation is that the different attention mechanisms (for different $k$'s) learn different weighting schemes over the source tokens that enable the prediction of the future token at time step $t+k$. Moreover, this is much more parameter efficient because the number of extra parameters introduced into the model is only $k \times d_{s_t} \times d_{s_t}$, where $d_{s_t} \ll |\mathcal{V}_{\mathrm{Target}}|$. Hence, we adopt the second implementation choice in our experiments.

\subsection{$K$-Step Ahead Unlikelihood Loss}
In Section~\ref{sec:future-mle}, we introduce an MLE-based loss for the task of $K$-step ahead token prediction. This idea can be naturally extended to the unlikelihood setting. Concretely, we impose the target and copy unlikelihood losses on the $K$-step ahead token prediction task as follows:

\begin{equation}
    \mathcal{L}_{K\mathrm{-StepTargetUL}}=-\sum_{t=1}^L \sum_{k=0}^K \gamma_k \sum_{c \in \mathcal{C}_{\mathrm{Target}}^{t+k}} \log \left(1- {P_{target}(c| \mathbf{y}_{1:t-1}, \mathbf{x}, \bm \theta)} \right)
    \label{eqn:kstep-tgt-ul}
\end{equation}

\begin{equation}
    \mathcal{L}_{K\mathrm{-StepCopyUL}}=-\sum_{t=1}^L \sum_{k=0}^K \gamma_k \sum_{c \in \mathcal{C}_{\mathrm{Copy}}^{t+k}} \log \left(1- {P_{copy}(c| \mathbf{y}_{1:t-1}, \mathbf{x}, \bm \theta)} \right)
    \label{eqn:kstep-copy-ul}
\end{equation}

\noindent where the negative candidate lists are $\mathcal{C}_{\mathrm{Target}}^{t+k} = \{y_1, \ldots , y_{t+k-1}\} \setminus \{y_{t+k}\}$ and $\mathcal{C}_{\mathrm{Copy}}^{t+k} = \{y_i\ \mid y_i \in \{y_1, \ldots , y_{t+k-1} \} \setminus \{y_{t+k}\} \enspace \mathrm{and} \enspace y_i \in \mathcal{V}_{\mathbf{x}} \}$. Penalizing the model for future repetitions through the $K$-step ahead unlikelihood losses should further enhance overall diversity of its outputs.

\subsection{Overall Training Objective}
To summarize, our S2S model is trained with a combination of likelihood and unlikelihood losses on the current ($k=0$) and future ($k=1,\ldots, K$) token prediction tasks.  The overall loss function is given by:

\begin{equation}
    \mathcal{L} = \mathcal{L}_{K\mathrm{-StepMLE}} + \lambda_T \mathcal{L}_{K\mathrm{-StepTargetUL}} + \lambda_C \mathcal{L}_{K\mathrm{-StepCopyUL}}
    \label{eqn:overall-loss}
\end{equation}
\noindent where $\lambda_T$ and $\lambda_C$ are hyperparameters that control the weight of target and copy UL losses respectively.

\section{Experiment Setup}
\label{sec:experiments}

\paragraph{Evaluation Metrics.}
\label{sec:eval-metrics}
To measure the quality of generated keyphrases, i.e., its relevance with respect to the source document, we compare the generated set of KPs to the KPs in the corresponding ground truth data. To this end, we report $F_1$@$M$, where $M$ refers to the number of model predicted keyphrases. We also include the corresponding precision and recall metrics. As justified in previous work \cite{reinforce-kp,yuan-catseqkp}, $F_1$@$M$ captures the ability of abstractive S2S models in generating a variable number of KPs depending on the source document, in comparison to traditional extractive methods where one is required to specify a cutoff in order to output the top-$k$ keyphrases. However, different from previous work, we report the overall $F_1$@$M$ score rather than separately computing this score for keyphrases present vs. absent in the source text. This is because our goal in this work is to overcome the lack of diversity issue in keyphrase generation models, and not necessarily to generate more absent keyphrases.

\noindent In order to evaluate the model outputs on the criterion of diversity, we define the following metrics:

\begin{addmargin}[1em]{0em}

$\%$ \textbf{Duplicate KPs} $=\left(1 -  \frac{\text{Number of Unique Keyphrases}}{\text{Total Number of Generated Keyphrases}}\right) * 100$
\vspace{0.3em}

\noindent$\%$ \textbf{Duplicate Tokens} $=\left(1 -  \frac{\text{Number of Unique Tokens}}{\text{Total Number of Generated Tokens}}\right) * 100$
\vspace{0.5em}

\noindent$\#$ \textbf{KPs}: We report the number of keyphrases generated. Ideally, the model should generate the same number of keyphrases as present in the ground truth target sequence. 
\end{addmargin}

\vspace{0.4em}

\noindent The next three metrics measure the inter-keyphrase similarity among the generated set of keyphrases - a lower value indicates fewer repetitions and thus more diversity in the output.

\begin{addmargin}[1em]{0em}
\vspace{0.4em}
\textbf{Self-BLEU}: We use Self-BLEU \cite{zhu2018texygen} which computes pairwise BLEU score \cite{papineni2002bleu} between generated KPs. This metric captures word level surface overlap. 
\vspace{0.3em}

\noindent \textbf{EditDist}\footnote{We utilize the \texttt{fuzzywuzzy} library in Python: \url{https://github.com/seatgeek/fuzzywuzzy}. It computes a score between 0 and 100, where 100 means exactly matching keyphrases.}: String matching can also be carried out at the character level. Through our EditDist metric, we calculate the pairwise Levenshtein Distance between KPs output by the model.
\vspace{0.3em}

\noindent \textbf{EmbSim}: With Self-BLEU and EditDist, we can only capture surface level repetitions between KPs. To overcome this limitation, we propose to use pre-trained phrase-level embeddings that measures inter-keyphrase similarity at a semantic level. Specifically, we compute pairwise cosine similarities between Sent2Vec embedding representations \cite{sent2vec} of keyphrases. Sent2Vec has been reported to perform well in previous  work on keyphrase extraction \cite{simple-unsupervised}.
\end{addmargin}
\vspace{0.3em}
All reported metrics are computed for each test set output, followed by averaging across all records. 

\paragraph{Datasets.}
\label{sec:datasets}
We carry out experiments on datasets from three domains \footnote{Additional results on five datasets that only provide a test set for evaluation (\texttt{INSPEC}, \texttt{KRAPIVIN}, \texttt{NUS}, \texttt{SEMEVAL}, \texttt{DUC}) are provided in the Appendix~\ref{sec:additional-results}.}: (1)
 \texttt{KP20K} \cite{meng-deep-kp} is a dataset of scientific articles; (2) \texttt{KPTimes} \cite{gallina-kptimes} consists of news articles and editor assigned keyphrases; (3) \texttt{StackEx} \cite{yuan-catseqkp} is a dataset curated from a community question answering forum with keyphrases being the user assigned tags.

\paragraph{Baselines.}
\label{sec:baselines}
We compare our approach to five S2S keyphrase generation baselines\footnote{We use the open-source code provided by \newcite{reinforce-kp} for implementing the baselines: \url{https://github.com/kenchan0226/keyphrase-generation-rl}} (4 MLE-based models and 1 which uses a reinforcement learning objective) --- (1) \textbf{catSeq}: A S2S model trained solely using the MLE objective. (refer to Equation~\ref{eqn:mle}). (2) \textbf{catSeqD}: Introduced by \newcite{yuan-catseqkp}, this method uses auxiliary semantic coverage and orthogonality losses to enhance generation diversity. (3)  \textbf{catSeqCorr}: \newcite{chen2018keyphrase} augment the attention scheme in the catSeq model with a coverage module and review mechanism. (4) \textbf{catSeqTG}: Instead of simply concatenating the article title and abstract together to form the source document, \newcite{tgnet-2018} design a model architecture that separately encodes the title information using an attention-guided matching layer. (5) \textbf{catSeqTG-2RF1}: \newcite{reinforce-kp} extend the catSeqTG model by training it using a reinforcement learning (RL) objective where $F_1$-score is directly used as the reward. Implementation details of our model are provided in Appendix~\ref{sec:additional-imp-details}.



\begin{table}[!t]
\small
  \centering
    \begin{tabular}{|c|l|ccc|cccccc}
    \cmidrule{2-11}    \multicolumn{1}{r}{} &       & \multicolumn{3}{c}{\textbf{Quality Evaluation}} \vrule & \multicolumn{6}{c}{\textbf{Diversity Evaluation}} \\
\cmidrule{2-11}    \multicolumn{1}{r}{} &       & \multicolumn{1}{P{2.2em}}{{$P@M$}} & \multicolumn{1}{P{2.2em}}{{$R@M$}} & \multicolumn{1}{P{2.5em}}{{$F_1$@$M$}} \vrule& \multicolumn{1}{p{2.4em}}{{$\#$KPs}} & \multicolumn{1}{P{4.5em}}{{$\%$Duplicate KPs $\downarrow$}} & \multicolumn{1}{P{4.5em}}{{$\%$Duplicate Tokens $\downarrow$}} & \multicolumn{1}{P{3.4em}}{{Self- BLEU  $\downarrow$}} & \multicolumn{1}{P{3em}}{{Edit- Dist  $\downarrow$}} & \multicolumn{1}{P{3em}}{{Emb- Sim  $\downarrow$}} \\
    \toprule
    \multirow{7}[8]{*}{\begin{sideways}Scientific Articles - \texttt{KP20K}\end{sideways}} & {Ground Truth} & -     & -     & -     & $\rightarrow$5.3   & 0.1   & 7.3   & 3.8   & 32.7  & 0.159 \\
\cmidrule{2-11}          & {catSeq} &    0.291   &   0.260    & 0.274 & 7.3   & 26.6  & 36.0  & 26.6  & 45.6  & 0.328 \\
          & {catSeqD} &   0.294    &    0.257   & 0.274 & 6.7   & 25.7  & 35.3  & 27.0  & 45.3  & 0.325 \\
          & {catSeqCorr} &   0.283    &    0.264   & 0.273 & 7.0   & 23.2  & 33.5  & 24.5  & 44.0  & 0.309 \\
          & {catSeqTG} &   \textbf{0.295}    &    0.262   & 0.278 & 6.8   & 24.7  & 34.3  & 26.2  & 45.2  & 0.323 \\
          & {catSeqTG-2RF1} &  0.274     &   \textbf{0.286}    & \textbf{0.280} & 7.5   & 30.9  & 41.7  & 30.7  & 46.7  & 0.341  \\
\cmidrule{2-11}          & {DivKGen (UL)} &   0.277    &    0.261   & 0.269 & {\textbf{5.0}} & 5.3   & 12.6  & 9.7   & \textbf{34.4} & \textbf{0.181} \\
          & {\quad $+$$K$-StepMLE} &   0.274    &   0.239    & 0.255 & 4.6   & 6.1   & 13.9  & 11.5  & 36.2  & 0.197 \\
          & {\quad $+$$K$-StepUL} &   0.273    & 0.240   & 0.256 & 4.6   & \textbf{4.9} & \textbf{11.7} & \textbf{8.8} & 35.2  & 0.185 \\ 
    \midrule \midrule \rule{0pt}{3ex} 
        \multirow{7}[8]{*}{\begin{sideways}News Articles - \texttt{KPTimes}\end{sideways}} & Ground Truth & -  & -  & -     & $\rightarrow$5.0   & 0.1   & 4.9   & 2.2   & 26.5  & 0.135 \\
\cmidrule{2-11}          & catSeq &  0.399     &   0.375    &  0.387   &    5.9   & 13.7      &   20.7    &    17.2   &    32.7   &  0.202 \\
          & catSeqD &   0.395    &  0.374     &   0.384    &   6.2    &  15.8     &   22.6    &    18.3   &   33.5    &  0.212 \\
          & catSeqCorr &   0.397    &   0.376    &    0.386   &   5.6   &   10.3   &   17.6    &    13.8   & 31.6   & 0.190 \\
          & catSeqTG &  \textbf{0.402}     & 0.380      &   \textbf{0.391}    &    5.9   & 13.8      &    21.2   &   17.6    &   32.8    &  0.203 \\
          & catSeqTG-2RF1 &   0.389    &    \textbf{0.386}   &   0.387    &   6.0	    &   14.0    &    21.0   & 18.6   &   32.5 &  0.192 \\
\cmidrule{2-11}          & DivKGen (UL) &   0.385    &   0.320    & 0.350 & {4.3} & \textbf{2.3}   & \textbf{7.0}   & \textbf{4.2}   & \textbf{27.8} & \textbf{0.142} \\
          & \quad $+$$K$-StepMLE &    0.391   &   0.316    & 0.349 & 4.3   & 3.3   & 7.9   & 5.3   & 27.9  & 0.147 \\
          & \quad $+$$K$-StepUL &    0.371   &   0.314    & 0.340 & \textbf{4.6}   & {3.6} & {8.7} & {5.8} & 28.3  & 0.149 \\
    \midrule \midrule \rule{0pt}{3ex} 
    \multirow{7}[8]{*}{\begin{sideways}Community QA- \texttt{StackEx}\end{sideways}} & Ground Truth & -     & -     & -     & $\rightarrow$2.7   & 0.3   & 2.9   & 1.5   & 24.2  & 0.167 \\
\cmidrule{2-11}          & catSeq &    0.526   &    0.518   &    0.522    &   \textbf{2.7}    &   4.3    &   7.4    &    4.1   &   28.2    & 0.226 \\
          & catSeqD &  0.510     &   0.524     &   0.517    &   2.8    &  5.0     &   8.6    &    4.8   &    28.8   &  0.230 \\
          & catSeqCorr &   0.501    &       0.526  &    0.513   &    2.9   &    5.4   &  9.3    &   5.2    &    29.1   & 0.235 \\
          & catSeqTG &    0.522   & 0.529      &   \textbf{0.526}    &   2.8    & 3.5      &   7.0    &    3.9   &    27.5   & 0.216 \\
          & catSeqTG-2RF1 &   0.433    &    \textbf{0.570}   &    0.492   &    3.8   &   6.7    &   11.8    &   6.2    &   29.0    & 0.220 \\
\cmidrule{2-11}          & DivKGen (UL) &  0.512     &   0.453    & 0.481 & 2.2 & \textbf{0.3}   & \textbf{1.4}   & \textbf{0.5}  & {23.3} & {0.175} \\
          & \quad $+$$K$-StepMLE &    \textbf{0.532}   &   0.438    & 0.480 & 2.0   & 0.4   & 1.5   & 0.6   & \textbf{23.1}  & 0.171 \\
          & \quad $+$$K$-StepUL &    0.516   &    0.454   & 0.483 & 2.2   & {0.4} & {1.6} & {0.7} & 23.7  & \textbf{0.170} \\
    \bottomrule
    \end{tabular}%
  \caption{KP generation results on datasets from 3 domains, evaluated on both quality and diversity criteria.} 
  \label{tab:main-results}
\end{table}

\section{Results and Analysis}

We report quality and diversity metrics on five baselines and three variants of the proposed approach 
(Table~\ref{tab:main-results}). We refer to our model as \textbf{DivKGen}; the base UL variant is trained with the regular MLE objective plus target and copy level unlikelihood losses. The rows denoted by $+K$-StepMLE and $+K$-StepUL are variants built on top of the base variant, by cumulatively incorporating $K$-Step ahead token prediction MLE and $K$-Step ahead UL losses respectively. For each dataset, we also report the ground truth statistics. For instance, the \texttt{KP20K} has an average keyphrase count of 5.3 in the test set with only 0.1\% duplicate KPs and 7.3\% duplicate tokens. In comparison, the MLE baseline (catSeq) produces a much higher percentage of repetitions. This is also evident from the inter-keyphrase pairwise similarity metrics Self-BLEU, EditDist and EmbSim. Surprisingly, the previous best performing model catSeqTG-2RF1, which uses RL to improve $F_1$ score, does worse than all MLE baselines in terms of diversity. 

In contrast, \textbf{DivKGen}, our proposed approach achieves much better diversity than all baselines. The repetition percentages are lowered and are relatively closer to the ground truth. 
There is a large boost by simply adding token and copy UL losses to the baseline MLE model. For \texttt{KP20K} dataset, we obtain small diversity gains through the incorporation of $K$-Step ahead losses whereas for the other two datasets, it does not result an improvement. A possible explanation is that the base DivKGen (UL) variant itself steers the diversity statistics to be quite close to that of the ground truth of these datasets. As a result, it becomes increasingly difficult to achieve a further reduction in this gap through any additional model changes. 

With regards to quality evaluation metrics, it can be observed that DivKGen models have slightly lower scores. 
This can be explained from a quality-diversity trade-off viewpoint. As the model attempts to explore the output space through the generation of more interesting KPs, it may output new KPs that are not present in the ground truth, thus resulting in lower precision. DivKGen generates  shorter sequences (and hence may not be able to produce all the KPs as per the ground truth) than the baselines, which could explain the lower recall.

\paragraph{Quality-Diversity Trade-off.}
We train different versions of DivKGen (UL) model on \texttt{KP20K} dataset by varying $\lambda_T$\footnote{For simplicity, we set $\lambda_T=\lambda_C$ to control the number of variable hyperparameters in the quality-diversity trade-off analysis.}, the UL loss coefficient (refer Equation~\ref{eqn:overall-loss}).
As depicted in Figure~\ref{fig:quality-div}, it can be seen that there is an obvious quality-diversity trade-off. For higher values of $\lambda_T$, we achieve a higher level of diversity (more unique KPs) at the cost of quality (and vice versa). Similar behaviour has been reported previously in the text generation literature \cite{bahuleyan-varattn,gao-jointly}. Hence, we recommend tuning the hyperparameters $\lambda_T$ and $\lambda_C$ to achieve a desired level of diversity. 

\paragraph{Ablation Study.}
We conduct an ablation analysis to investigate the effect of different losses. We start with the MLE baseline and add loss components one-by-one as presented in Table~\ref{tab:ablation}. It is evident that the best diversity is obtained while using the full model (last row). Also, interestingly each individual loss component by itself (i.e., TargetUL, CopyUL and $K$-StepMLE), is not as effective as their combination. This suggests that each of the losses contribute in a synergetic manner to maximize diversity gains. 

\begin{table}[!t]
\begin{minipage}[b]{0.56\linewidth}
\small
  \centering
    \begin{tabular}{p{9.5em}|c|c|c|c}
    \toprule
    \textbf{DivKGen Variants} & \multicolumn{1}{P{4.0em}|}{{Overall $F_1$@$M$ $\uparrow$}} & \multicolumn{1}{P{4.1em}|}{{\%Duplicate KPs $\downarrow$}} & \multicolumn{1}{P{4.1em}|}{{\%Duplicate Tokens $\downarrow$}} & \multicolumn{1}{P{3em}}{{Self-BLEU$\downarrow$}} \\
    \midrule
    w/ TargetUL & \textbf{0.277} & 12.0 & 19.8 & 16.7 \\
    w/ CopyUL & 0.263 & 14.1  & 22.7  & 19.9 \\
    w/ $K$-StepMLE & 0.265 & 12.6  & 18.9  & 16.3 \\
    w/ TargetUL $+$ CopyUL & 0.269 & 5.3   & 12.6  & 9.7 \\
    \quad  $+$ $K$-StepMLE & 0.255 & 6.1   & 13.9  & 11.5 \\
    \quad  $+$ $K$-StepUL & 0.256 & \textbf{4.9}   & \textbf{11.7}  & \textbf{8.8} \\
    \bottomrule
    \end{tabular}
  \caption{Ablation study on the \texttt{KP20k} dataset. Each row denotes a DivKGen model variant obtained by adding the specified component. The last row corresponds to the full model.}
  \label{tab:ablation}
\end{minipage}\hfill
\begin{minipage}[b]{0.36\linewidth}
    \small
    \includegraphics[width=1.1\textwidth]{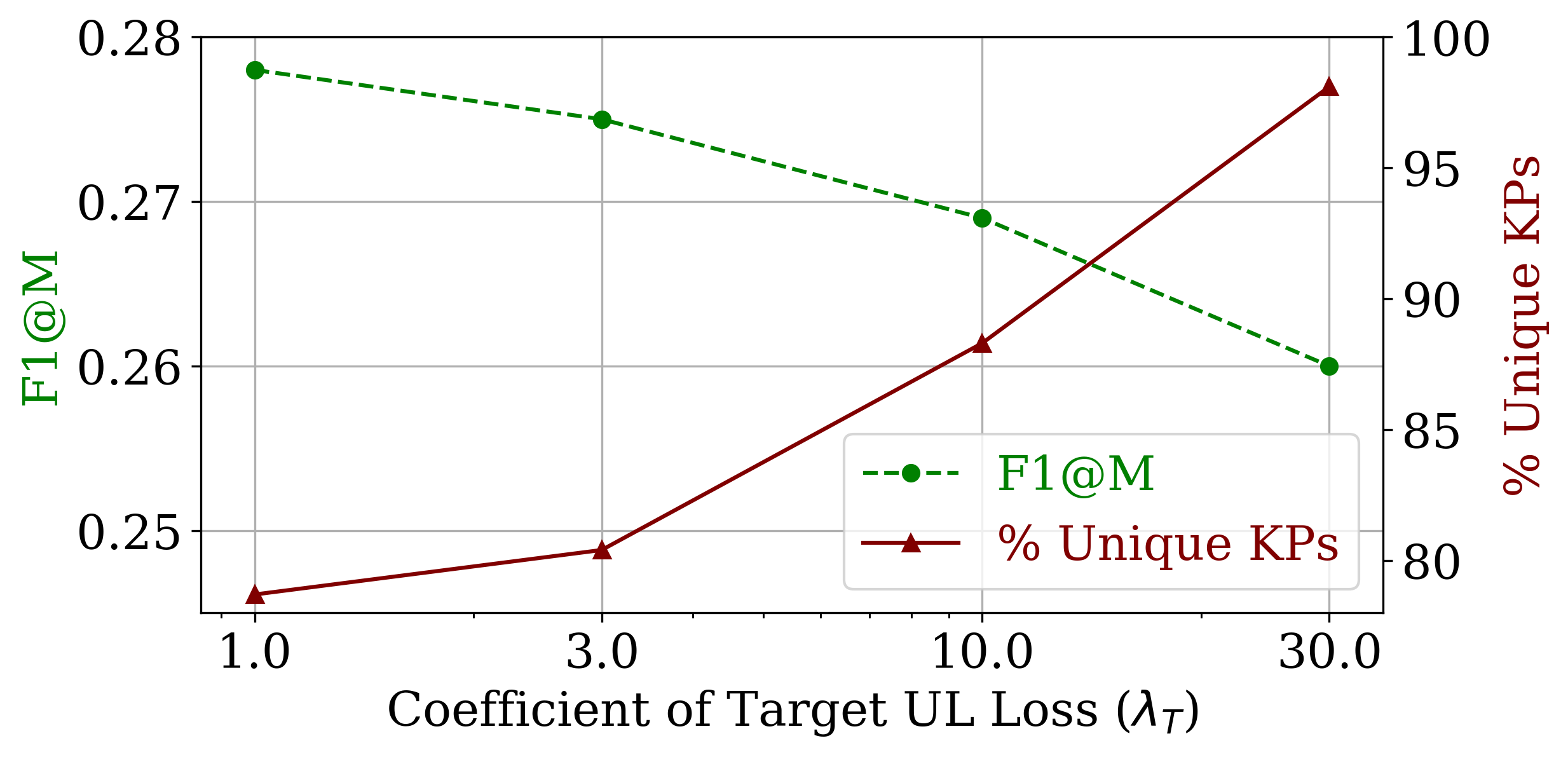}
    \captionof{figure}{Illustration of quality-diversity trade-off: \%Unique KPs = $(100 - $\%Duplicate KPs$)$ is used as a representative metric for diversity. }
    \label{fig:quality-div}
\end{minipage}
\end{table}


\section{Related Work}
\paragraph{Keyphrase Generation and Extraction.}
Traditional KP extraction methods such as TextRank \cite{textrank} and TopicRank \cite{topicrank} first select candidate phrases from the source document using heuristics and then rank these candidates based on some measure of relevance or importance.
\newcite{meng-deep-kp} formulate keyphrase generation as a S2S learning problem, with an advantage over previous extractive methods that it could even generate relevant KPs absent from the source text. A limitation of their approach is that one was still required to rank the top-$k$ KPs. This was addressed in works by \cite{semi-supervised-kp} and \cite{yuan-catseqkp} which could generate a variable number of KPs depending on the input. We adopt a similar setup but carry out a comprehensive analysis of such models in terms of their output diversity, which has been largely ignored in previous work. 

\paragraph{Diversity in Language Generation.}
Diversity promoting objectives for text generation have been previously explored in the literature \cite{li-diversity,avgout,tldr}. However, these studies examine the overall corpus level diversity. For instance, the lack of diversity in a dialogue system, due to the fact that the model generates frequently seen responses from the training set. In our case, we address a different kind of diversity issue, arising as a result of repetitions occurring within individual outputs. Thus neural unlikelihood training \cite{welleck-ul} is well suited to our problem. 
Test time decoding strategies to improve diversity such as top-$k$ sampling \cite{topk-sampling}, nucleus sampling  \cite{nucleus-sampling} and diverse beam search \cite{diverse-beam-search} are orthogonal to our approach and can naturally be incorporated.

\section{Conclusion and Future Work}
In this work, we first point out the shortcomings of MLE based training for keyphrase generation. We specifically address the lack of output diversity issue via the use of unlikelihood training objective. We adopt a target level unlikelihood loss and propose a novel copy token unlikelihood loss, the combination of which provides large diversity gains. In addition, a $K$-step ahead MLE and UL objective is incorporated into the training. Through extensive experiments on datasets from three different domains, we demonstrate the effectiveness of our model for diverse keyphrase generation. For future work, we plan to explore directions that would enable us to simultaneously optimize for quality and diversity metrics.

\section*{Acknowledgements}
We thank Francis Duplessis, Ivan Kobyzev, Jackie Chi Kit Cheung, Kushal Arora, Marjan Albooyeh, Mehran Kazemi, Simon Prince and Wenjie Zi for valuable inputs and helpful discussions.


\bibliographystyle{coling}
\bibliography{references}

\newpage
\appendix

\onecolumn
\section*{\centering{\Large{ Appendix}}}
\section*{\centering{\Large{ Diverse Keyphrase Generation with Neural Unlikelihood Training}}}

\section{Dataset Statistics}
\label{sec:dataset-stats}
Table 5 presents the number of instances in each dataset, split across training, validatation and test sets. 

\begin{table}[!h]
\small
  \centering
    \begin{tabular}{P{9.335em}|c|c|c}
    \toprule
    \multicolumn{1}{P{9.335em}|}{\textbf{Dataset}} & \multicolumn{1}{P{5em}|}{\textbf{$\#$Train}} & \multicolumn{1}{P{5em}|}{\textbf{$\#$Validation}} & \multicolumn{1}{P{5em}}{\textbf{$\#$Test}} \\
    \midrule
    \texttt{KP20K} & 530k  & 20k   & 20k \\
    \texttt{KPTimes}* & 260k  & 10k   & 20k \\
    \texttt{StackEx} & 299k  & 16k   & 16k \\
    \midrule
    \texttt{INSPEC} &   $-$    &  1500   &  500 \\
    \texttt{SEMEVAL} &    $-$   &   144   &  100 \\
    \texttt{KRAPIVIN} &   $-$    &    1844   &  460 \\
    \texttt{NUS}   &   $-$    &    $-$   &  211 \\
    \texttt{DUC}   &    $-$   &    $-$   &  308 \\
    \bottomrule
    \end{tabular}%
  \label{tab:dataset-stats}%
  \caption{Train/validation/test statistics of the datasets used in this work.}
\end{table}%
\noindent *Note that our test set for \texttt{KPTimes} is a combination of 10k records from \texttt{KPTimes} and 10k records from \texttt{JPTimes} \cite{gallina-kptimes}.

\section{Implementation Details}
\label{sec:additional-imp-details}
We use the AllenNLP package \cite{allennlp}, which is built on PyTorch framework \cite{pytorch}, for implementing our models.
We provide as input to the model the concatenated title and abstract. Following \cite{yuan-catseqkp}, the ground truth target keyphrases are arranged as a sequence, where the absent KPs follow the present KPs. The size of source and target vocabularies are set to $50$k and $10$k respectively. The delimiter token that is inserted in between target keyphrases is denoted as \texttt{<SEP>}. Both the LSTM\footnote{Our initial experiments with Transformer-based architectures \cite{vaswani2017attention} showed similar performance, but required a lot more parameters. We thus carry out experiments with LSTM models as they are more parameter efficient and had lesser computational requirements. A similar observation has been reported in \cite{order-matter}} encoder and decoder have a hidden size of $100$d. Word embeddings on both the source and target side are also set to $100$d and randomly initialized. We use Adam optimizer \cite{adam} with default parameters to train the model. The batch size is set to 64 and we incorporate early stopping based on validation $F_1$ score as the criterion. 

Regarding the loss term coefficients for UL losses and $K$-step ahead loss, we set $\lambda_T=15.0$, $\lambda_C=18.0$ and $\gamma_0=1.0$, which are obtained based on performance on validation set after grid search hyperparameter optimization. The hyperparameter tuning is carried out on \texttt{KP20K} dataset and the best values are adopted for other datasets too. The value of $K$ is set to 2, which corresponds to upto $2$-step ahead prediction.

For test time decoding, unlike previous work \cite{semi-supervised-kp,chen-integrated-kp,yuan-catseqkp}, we do not apply exhaustive decoding with large beam sizes, followed by pruning and de-duplication of the output. This is because our model is trained to generate outputs without repetitions. As such, we do not require any adhoc post-processing strategies to improve diversity. Thus we adopt greedy decoding at test time as well, similar to \cite{reinforce-kp}. For quality evaluation, we use the evaluation scripts\footnote{\url{https://github.com/kenchan0226/keyphrase-generation-rl}} provided by \cite{reinforce-kp}. Note that Porter Stemming is applied on the outputs for the purpose of quality evaluation.

\section{Results on Evaluation-Only Datasets}
\label{sec:additional-results}
We present additional results in Table~\ref{tab:additional-results-1} and Table~\ref{tab:additional-results-2} on the following datasets: \texttt{INSPEC} \cite{inspec}, \texttt{KRAPIVIN} \cite{krapivin}, \texttt{NUS} \cite{nus}, \texttt{SEMEVAL} \cite{semeval} and \texttt{DUC} \cite{duc}. These datasets are smaller in size and hence, similar to previous work we only use them as test sets.  \texttt{DUC} is a dataset with news articles and their associated keyphrases. Hence, we use the models trained on \texttt{KPTimes} for evaluation on \texttt{DUC}. Since the remaining datasets are from the domain of scientific articles, we test them using the best checkpoints obtained from training on \texttt{KP20K} dataset. 

\begin{table}[!ht]
\small
  \centering
    \begin{tabular}{|c|l|ccc|cccccc}
    \cmidrule{2-11}    \multicolumn{1}{r}{} &       & \multicolumn{3}{c}{\textbf{Quality Evaluation}} \vrule & \multicolumn{6}{c}{\textbf{Diversity Evaluation}} \\
\cmidrule{2-11}    \multicolumn{1}{r}{} &       & \multicolumn{1}{P{2.2em}}{{$P@M$}} & \multicolumn{1}{P{2.2em}}{{$R@M$}} & \multicolumn{1}{P{2.5em}}{{$F_1$@$M$}} \vrule& \multicolumn{1}{p{2.4em}}{{$\#$KPs}} & \multicolumn{1}{P{4.5em}}{{$\%$Duplicate KPs $\downarrow$}} & \multicolumn{1}{P{4.5em}}{{$\%$Duplicate Tokens $\downarrow$}} & \multicolumn{1}{P{3.4em}}{{Self- BLEU  $\downarrow$}} & \multicolumn{1}{P{3em}}{{Edit- Dist  $\downarrow$}} & \multicolumn{1}{P{3em}}{{Emb- Sim  $\downarrow$}} \\
    \toprule
    \multirow{7}[8]{*}{\begin{sideways} \texttt{SEMEVAL}\end{sideways}} & {Ground Truth} & - & - & - & $\rightarrow$15.1 & 1.6 & 26.6 & 12.7 & 32.6 & 0.152 \\
\cmidrule{2-11}          & {catSeq} &    0.321 & 0.105 & 0.158 & \textbf{12.1} & 46.2 & 53.8 & 31.9 & 52.3 & 0.415 \\
          & {catSeqD} &   0.306 & 0.105 & 0.157 & 11.5 & 43.3 & 53.3 & 33.2 & 53.5 & 0.420 \\
          & {catSeqCorr} &   0.291 & 0.102 & 0.151 & 9.5 & 29.9 & 39.8 & 24.2 & 45.7 & 0.322 \\
          & {catSeqTG} &  0.325 & 0.099 & 0.152 & 11.8 & 45.2 & 53.5 & 34.0 & 55.5 & 0.450 \\
          & {catSeqTG-2RF1} & 0.338 & 0.117 & 0.174 & 7.5 & 32.0 & 41.3 & 29.7 & 46.5 & 0.327  \\
\cmidrule{2-11}          & {DivKGen (UL)} &   \textbf{0.341} & \textbf{0.155} & \textbf{0.213} & 4.8 & 4.8 & 13.1 & 8.4 & 36.0 & 0.177 \\
          & {\quad $+$$K$-StepMLE} &   0.340 & 0.142 & 0.201 & 4.4 & \textbf{4.3} & 14.7 & 10.2 & 37.6 & 0.194 \\
          & {\quad $+$$K$-StepUL} &  0.339 & 0.135 & 0.193 & 4.4 & 4.6 & \textbf{10.9} & \textbf{6.9} & \textbf{35.2} & \textbf{0.171} \\ 
    \midrule \midrule \rule{0pt}{3ex} 
        \multirow{7}[8]{*}{\begin{sideways} \texttt{INSPEC}\end{sideways}} & {Ground Truth} & -     & -     & -     & $\rightarrow$9.8   & 0.3 & 15.7 & 7.6 & 33.8 & 0.168 \\
\cmidrule{2-11}          & {catSeq} &    0.301 & 0.161 & 0.210 & 10.8 & 39.4 & 49.3 & 29.6 & 50.4 & 0.396 \\
          & {catSeqD} &   0.289 & 0.146 & 0.194 & 9.5 & 36.1 & 46.4 & 29.5 & 48.6 & 0.376 \\
          & {catSeqCorr} &   0.281 & 0.153 & 0.198 & \textbf{9.8} & 33.6 & 43.7 & 26.4 & 47.0 & 0.351 \\
          & {catSeqTG} &  0.308 & 0.163 & 0.213 & 11.6 & 41.3 & 51.6 & 31.6 & 51.0 & 0.405 \\
          & {catSeqTG-2RF1} & 0.302 & 0.165 & 0.213 & 7.9 & 37.6 & 47.7 & 32.1 & 51.5 & 0.402  \\
\cmidrule{2-11}          & {DivKGen (UL)} &   \textbf{0.375} & \textbf{0.226} & \textbf{0.282} & 5.1 & \textbf{6.2} & \textbf{13.5} & \textbf{11.3} & \textbf{33.3} & \textbf{0.172}\\
          & {\quad $+$$K$-StepMLE} &   0.366 & 0.207 & 0.264 & 4.8 & 7.5 & 15.7 & 13.6 & 35.9 & 0.194 \\
          & {\quad $+$$K$-StepUL} &  0.360 & 0.200 & 0.257 & 4.9 & 6.8 & 14.0 & 11.5 & 35.7 & 0.176 \\ 
    \midrule \midrule \rule{0pt}{3ex} 
        \multirow{7}[8]{*}{\begin{sideways} \texttt{KRAPIVIN}\end{sideways}} & {Ground Truth} & - & - & - & $\rightarrow$5.7 & 0.1 & 9.8 & 4.6 & 34.6 & 0.174 \\
\cmidrule{2-11}          & {catSeq} &    \textbf{0.289} & 0.247 & \textbf{0.266} & 8.4 & 33.5 & 42.5 & 28.3 & 49.8 & 0.381 \\
          & {catSeqD} &   0.280 & 0.234 & 0.255 & 7.3 & 29.4 & 39.6 & 27.5 & 48.2 & 0.358 \\
          & {catSeqCorr} &   0.264 & 0.237 & 0.249 & 8.4 & 30.2 & 39.7 & 26.1 & 46.6 & 0.346 \\
          & {catSeqTG} &  0.267 & 0.235 & 0.250 & 8.2 & 30.2 & 40.3 & 28.0 & 48.3 & 0.362 \\
          & {catSeqTG-2RF1} & 0.273 & \textbf{0.257} & 0.265 & 7.4 & 32.3 & 42.2 & 29.7 & 47.8 & 0.357  \\
\cmidrule{2-11}          & {DivKGen (UL)} &   0.244 & 0.237 & 0.240 & \textbf{5.8} & \textbf{6.7} & \textbf{14.2} & \textbf{9.2} & \textbf{34.0} & \textbf{0.182} \\
          & {\quad $+$$K$-StepMLE} & 0.263 & 0.221 & 0.241 & 5.1 & 8.1 & 15.8 & 11.9 & 36.8 & 0.209 \\
          & {\quad $+$$K$-StepUL} &  0.258 & 0.227 & 0.242 & 5.5 & 8.4 & 15.0 & 10.5 & 35.7 & 0.194 \\ 
    \bottomrule
    \end{tabular}%
  \caption{Results of keyphrase generation on \texttt{SEMEVAL}, \texttt{INSPEC} and \texttt{KRAPIVIN} datasets.} 
  \label{tab:additional-results-1}
\end{table} 

\begin{table}[!h]
\small
  \centering
    \begin{tabular}{|c|l|ccc|cccccc}
    \cmidrule{2-11}    \multicolumn{1}{r}{} &       & \multicolumn{3}{c}{\textbf{Quality Evaluation}} \vrule & \multicolumn{6}{c}{\textbf{Diversity Evaluation}} \\
\cmidrule{2-11}    \multicolumn{1}{r}{} &       & \multicolumn{1}{P{2.2em}}{{$P@M$}} & \multicolumn{1}{P{2.2em}}{{$R@M$}} & \multicolumn{1}{P{2.5em}}{{$F_1$@$M$}} \vrule& \multicolumn{1}{p{2.4em}}{{$\#$KPs}} & \multicolumn{1}{P{4.5em}}{{$\%$Duplicate KPs $\downarrow$}} & \multicolumn{1}{P{4.5em}}{{$\%$Duplicate Tokens $\downarrow$}} & \multicolumn{1}{P{3.4em}}{{Self- BLEU  $\downarrow$}} & \multicolumn{1}{P{3em}}{{Edit- Dist  $\downarrow$}} & \multicolumn{1}{P{3em}}{{Emb- Sim  $\downarrow$}} \\
    \toprule
        \multirow{7}[8]{*}{\begin{sideways} \texttt{NUS}\end{sideways}} & {Ground Truth} & - & - & - & $\rightarrow$11.7 & 5.3 & 23.6 & 12.3 & 32.8 & 0.161 \\
\cmidrule{2-11}          & {catSeq} &    0.391 & 0.210 & 0.274 & \textbf{11.7} & 43.6 & 52.0 & 31.6 & 53.7 & 0.442 \\
          & {catSeqD} &   0.397 & 0.206 & 0.271 & 10.4 & 41.4 & 49.6 & 32.2 & 52.9 & 0.433 \\
          & {catSeqCorr} &   0.396 & 0.217 & 0.281 & 10.7 & 38.9 & 47.8 & 29.8 & 50.1 & 0.398 \\
          & {catSeqTG} &  \textbf{0.407} & 0.203 & 0.271 & 11.3 & 42.9 & 51.8 & 33.6 & 54.3 & 0.445 \\
          & {catSeqTG-2RF1} & 0.385 & 0.228 & 0.286 & 7.6 & 32.6 & 44.1 & 30.0 & 47.4 & 0.355  \\
\cmidrule{2-11}          & {DivKGen (UL)} &   0.376 & \textbf{0.238} & \textbf{0.292} & 5.3 & 6.5 & 15.0 & 10.3 & \textbf{34.8} & \textbf{0.189} \\
          & {\quad $+$$K$-StepMLE} &   0.394 & 0.225 & 0.287 & 4.8 & 8.6 & 17.7 & 14.1 & 37.7 & 0.218 \\
          & {\quad $+$$K$-StepUL} &  0.393 & 0.218 & 0.281 & 4.4 & \textbf{5.9} & \textbf{13.5} & \textbf{10.0} & 36.6 & 0.202 \\ 
    \midrule \midrule \rule{0pt}{3ex} 
    \multirow{7}[8]{*}{\begin{sideways}\texttt{DUC}\end{sideways}} & Ground Truth & - & - & - & $\rightarrow$8.1 & 0.2 & 14.1 & 6.4 & 33.4 & 0.176 \\
\cmidrule{2-11}          & catSeq &    0.106 & 0.059 & 0.076 & 5.9 & 19.5 & 28.5 & 24.6 & 38.0 & 0.243 \\
          & catSeqD &  0.104 & 0.057 & 0.074 & 6.2 & 20.5 & 29.8 & 24.8 & 38.1 & 0.249 \\
          & catSeqCorr &   0.103 & 0.057 & 0.073 & 5.5 & 15.0 & 24.9 & 20.3 & 36.8 & 0.226 \\
          & catSeqTG &    0.111 & 0.060 & 0.078 & 5.7 & 18.0 & 27.8 & 22.8 & 37.1 & 0.231 \\
          & catSeqTG-2RF1 &   0.115 & 0.069 & 0.086 & \textbf{6.2} & 19.4 & 28.9 & 27.1 & 36.2 & 0.217 \\
\cmidrule{2-11}          & DivKGen (UL) &  0.135 & 0.065 & 0.088 & 4.2 & 3.4 & \textbf{9.5} & \textbf{5.6} & 30.4 & 0.151 \\
          & \quad $+K$-StepMLE &    \textbf{0.152} & 0.069 & \textbf{0.095} & 4.0 & \textbf{3.0} & 9.7 & 5.7 & 30.8 & 0.148 \\
          & \quad $+K$-StepUL &  0.143 & \textbf{0.070} & 0.094 & 4.5 & 3.8 & 9.9 & 6.7 & \textbf{29.4} & \textbf{0.139} \\
    \bottomrule
    \end{tabular}%
    \caption{Results of keyphrase generation on \texttt{NUS} and \texttt{DUC} datasets.}
  \label{tab:additional-results-2}
\end{table}

\newpage
\section{Qualitative Results}
\label{sec:qualitative-results}
In Tables 8, 9 and 10, we present qualitative results from the three domains respectively, i.e., scientific articles, news and community QA forums. The input to each model is the title and the abstract, and the expected output is displayed as the ground truth. In these case study examples, it can be observed that both the MLE and RL baseline tend to generate numerous repetitions in their output sequence. Our DivKGen base variant (UL) achieves good diversity, although occasionally it does generate few repetitions. However, we are able to avoid duplicates with the DivKGen (Full) model, which additionally incorporates the $K$-step ahead losses. We attribute this to be due to the enhanced model planning capabilities that DivKGen (Full) exhibits, by learning what the future tokens should or shouldn't be. 

\begin{table}[!h]
\small
  \centering
    \begin{tabular}{|p{9.915em}|p{35.25em}|}
    \toprule
    \multicolumn{2}{|P{45.165em}|}{\textbf{Dataset : \texttt{KP20K}}} \\
    \midrule
    \textbf{Title} & automatic image segmentation by dynamic region merging . \\
    \midrule
    \textbf{Abstract} & this paper addresses the automatic image segmentation problem in a region merging style . with an initially oversegmented image , in which many regions or superpixels with homogeneous color are detected , an image segmentation is performed by iteratively merging the regions according to a statistical test . there are two essential issues in a region merging algorithm order of merging and the stopping criterion . in the proposed algorithm , these two issues are solved by a novel predicate , which is defined by the sequential probability ratio test and the minimal cost criterion . starting from an oversegmented image , neighboring regions are progressively merged if there is an evidence for merging according to this predicate . we show that the merging order follows the principle of dynamic programming . this formulates the image segmentation as an inference problem , where the final segmentation is established based on the observed image . we also prove that the produced segmentation satisfies certain global properties . in addition , a faster algorithm is developed to accelerate the region merging process , which maintains a nearest neighbor graph in each iteration . experiments on real natural images are conducted to demonstrate the performance of the proposed dynamic region merging algorithm . \\
    \midrule
    \textbf{Ground Truth} & \textit{image segmentation ; region merging ; dynamic programming ; wald sequential probability ratio test} \\
    \midrule\midrule
    \textbf{catSeq MLE Baseline} & \textit{image segmentation ; region merging ; \textcolor{red}{region merging} ; dynamic programming ; \textcolor{red}{image segmentation}} \\
    \midrule
    \textbf{catSeqTG-2RF1 (RL)} & \textit{image segmentation ; region merging ; dynamic programming ; \textcolor{red}{image segmentation} ; \textcolor{red}{dynamic programming}} \\
    \midrule
    \textbf{DivKGen (UL)} & \textit{image segmentation ; region merging ; \textcolor{red}{region merging} ; dynamic programming ; nearest neighbor graph} \\
    \midrule
    \textbf{DivKGen (Full)} & \textit{image segmentation ; dynamic programming ; region merging ; stopping criterion} \\
    \bottomrule
    \end{tabular}%
  \label{tab:qualitative-kp20k}
  \caption{Case Study on \texttt{KP20K} dataset.}
\end{table}%

\begin{table}[!h]
\small
  \centering
    \begin{tabular}{|p{9.915em}|p{35.25em}|}
    \toprule
    \multicolumn{2}{|P{45.165em}|}{\textbf{Dataset : \texttt{KPTimes}}} \\
    \midrule
    \textbf{Title} & n.f.l . said to be closer to testing for h.g.h .  \\
    \midrule
    \textbf{Abstract} & the n.f.l . owners and players have figured out how to divide up their money , and have spent a busy week reconstituting rosters and renewing rivalries . but there is still unfinished business in their labor standoff , and the most important issue remaining could be the question of drug testing . the n.f.l . , whose new collective bargaining agreement is expected to be completed and ratified by thursday , could begin blood testing for human growth hormone as soon as september , according to a person briefed on the negotiations who was not authorized to speak publicly , making it the first major north american sports league to conduct such testing on its top players with the union consent . players had long resisted blood testing under the former union president gene upshaw , and negotiators are still determining ways to make the program acceptable to current players . details to be worked out include how many players will be tested for performance enhancing drugs and how they would be randomly selected when drug testing resumes . there was no drug testing of any kind conducted during the lockout . but commissioner roger goodell and demaurice smith , the players union executive director , were said by people briefed on negotiations to have long seen the need for growth hormone testing and to want to cast the n.f.l . as a leader in combating drugs in major sports . they have pointed to the joint actions of upshaw and the former commissioner paul tagliabue , who moved to start the steroid testing program in the late . i think both sides have a commitment to being leaders in this area and to having the best \\
    \midrule
    \textbf{Ground Truth} & \textit{human growth hormone ; goodell roger ; national football league ; doping sports ; football ; organized labor ; smith demaurice ; tests drug use} \\
    \midrule \midrule
    \textbf{catSeq MLE Baseline} & \textit{human growth hormone ; national football league ; football ; tests and testing ; \textcolor{red}{national football league} ; tests drug use ; \textcolor{red}{tests drug use} ; \textcolor{red}{national football league} ; \textcolor{red}{tests drug use} ; doping sports ; \textcolor{red}{tests drug use} ; \textcolor{red}{national football league} ; \textcolor{red}{tests drug use}} \\
    \midrule
    \textbf{catSeqTG-2RF1 (RL)} & \textit{human growth hormone ; baseball ; national football league ; tests drug use ; \textcolor{red}{national football league} ; football ; \textcolor{red}{national football league} ; lockouts ; organized labor} \\
    \midrule
    \textbf{DivKGen (UL)} & \textit{human growth hormone ; drug abuse and traffic ; national football league ; goodell roger ; lockouts ; \textcolor{red}{national football league}} \\
    \midrule
    \textbf{DivKGen (Full)} & \textit{human growth hormone ; upshaw gene ; goodell roger ; national football league ; organized labor ; lockouts ; football} \\
    \bottomrule
    \end{tabular}%
  \label{tab:qualitative-kptimes}
  \caption{Case Study on \texttt{KPTimes} dataset.}
\end{table}%

\begin{table}[!h]
\small
  \centering
    \begin{tabular}{|p{9.915em}|p{35.25em}|}
    \toprule
    \multicolumn{2}{|P{45.165em}|}{\textbf{Dataset : \texttt{StackEx}}} \\
    \midrule
    \textbf{Title} & do deep learning algorithms represent ensemble based methods ? \\
    \midrule
    \textbf{Abstract} & shortly about deep learning for reference ) : deep learning is a branch of machine learning based on a set of algorithms that attempt to model high level abstractions in data by using a deep graph with multiple processing layers , composed of multiple linear and non linear transformations.various deep learning architectures such as deep neural networks , convolutional deep neural networks , deep belief networks and recurrent neural networks have been applied to fields like computer vision , automatic speech recognition , natural language processing , audio recognition and bioinformatics where they have been shown to produce state of the art results on various tasks.my question can deep neural networks or convolutional deep neural networks be viewed as ensemble based method of machine learning or it is different approaches \\
    \midrule
    \textbf{Ground Truth} & \textit{deep learning ; machine learning ; neural networks ; convolutional neural networks} \\
    \midrule \midrule
    \textbf{catSeq MLE Baseline} & \textit{deep learning ; machine learning ; \textcolor{red}{deep learning}} \\
    \midrule
    \textbf{catSeqTG-2RF1 (RL)} & \textit{deep learning ; machine learning ; neural network ; \textcolor{red}{machine learning}} \\
    \midrule
    \textbf{DivKGen (UL)} & \textit{deep learning ; machine learning ; ensemble modeling} \\
    \midrule
    \textbf{DivKGen (Full)} & \textit{deep learning ; neural networks} \\
    \bottomrule
    \end{tabular}%
  \label{tab:qualitative-stackex}
  \caption{Case Study on \texttt{StackEx} dataset.}
\end{table}%

\end{document}